\documentclass[conference,9pt,letterpaper]{IEEEtran}

\newif\ifanonymous
\anonymousfalse 
\newif\ifaidisclosure
\aidisclosuretrue

\usepackage[T1]{fontenc}
\usepackage{cite}
\usepackage{amsmath,amssymb}
\usepackage{algorithmic}
\usepackage{graphicx}
\usepackage{xcolor}
\usepackage{flafter} 
\usepackage{booktabs}
\usepackage{balance} 
\usepackage[hidelinks]{hyperref}
\graphicspath{{fig/}}

\hypersetup{pdfauthor={},pdfsubject={},pdfkeywords={},pdfcreator={}}
\makeatletter
\AtBeginDocument{\hypersetup{pdftitle={\@title},pdfauthor={}}}
\makeatother

\newcounter{dacshippedpages}

\AddToHook{shipout/after}{\stepcounter{dacshippedpages}}

\title{Selective ROM Weight Storage for Responsive LLM Control on Edge Robots}

\ifanonymous
  \author{}
\else

\author{
  \IEEEauthorblockN{
    Boxiang Song\IEEEauthorrefmark{1},
    Zhihong Chen\IEEEauthorrefmark{2},
    Xiaoyu Hao\IEEEauthorrefmark{1},
    Jinhong Huang\IEEEauthorrefmark{1},
    Chengkai Zhu\IEEEauthorrefmark{1},\\
    Jiaqi Yang\IEEEauthorrefmark{3},
    Hongyang Pan\IEEEauthorrefmark{4},
    Changhao Yan\IEEEauthorrefmark{1},
    and Keren Zhu\IEEEauthorrefmark{1}
  }
  \IEEEauthorblockA{
    \IEEEauthorrefmark{1}Fudan University, Shanghai, China\\
    \IEEEauthorrefmark{2}Shanghai Jiao Tong University, Shanghai, China\\
    \IEEEauthorrefmark{3}Peking University, Beijing, China\\
    \IEEEauthorrefmark{4}Shanghai Trace Intelligence Co., Ltd. Shanghai, China
  }
  \IEEEauthorblockA{
    bxsong25@m.fudan.edu.cn, mee3210@sjtu.edu.cn, xyhao26@m.fudan.edu.cn,\\
    huangjh24@m.fudan.edu.cn, ckzhu25@m.fudan.edu.cn, 25001111916@stu.pku.edu.cn,\\
    panhongyang@traceic.com, yanch@fudan.edu.cn, krzhu@fudan.edu.cn
  }
}

\fi

\begin{document}
\bstctlcite{referenceAuthorControl}
\maketitle

\begin{abstract}
Edge robots increasingly use large language models (LLMs) for complex tasks, demanding efficient inference alongside timely control. Limited memory capacity constrains model deployment, while repeated weight transfers during decoding compete with activations and KV cache for external bandwidth. Sequential generation of robot actions further delays decisions and periodic control on shared hardware. We present a hybrid memory architecture with heterogeneous scheduling to address these bottlenecks jointly. Banked ROM stores selected fixed weights and supplies a standard compute array through a dedicated datapath. LPDDR retains the remaining weights and mutable state, preserving model flexibility. Coordinated operand delivery and preemption at output slice boundaries accelerate inference while limiting interference with periodic control. A structured decision interface scores candidate phases in a single forward pass, avoiding sequential token generation for routine choices. Together, these mechanisms improve memory service, compute access, and decision execution for edge robotic workloads. Experiments with Qwen3-8B achieve a peak decode speedup of 1.56$\times$ and reduce P99 task response time by a factor of approximately 227.27$\times$, while completing 100.00\% of the tested grasping and transport tasks.

\end{abstract}

\begin{IEEEkeywords}
  Edge robotics, LLM inference, memory architecture, scheduling algorithm, hardware-software co-design. 
\end{IEEEkeywords}

\section{Introduction}
\label{sec:introduction}
Large language models (LLMs) allow robots to map instructions to physical skills~\cite{saycan}, generate control code~\cite{cap}, and convert vision-language inputs into actions~\cite{rt2,openvla}. Cloud offloading adds network delays and risks stale plans~\cite{fogros}, so dynamic environments require running these billion-parameter models locally at the edge. However, local execution faces a severe memory bottleneck~\cite{roofline,llmcompass,neupims}: limited DRAM capacity restricts model size, and heavy traffic from weights, activations, and KV cache slows decoding. Furthermore, running high-level planning alongside periodic control on shared hardware~\cite{prema,camdn,planaria,neu10} requires both low decision latency~\cite{roborun} and strict control deadlines.

\begin{figure}[!t]
\centering
\includegraphics[width=\columnwidth]{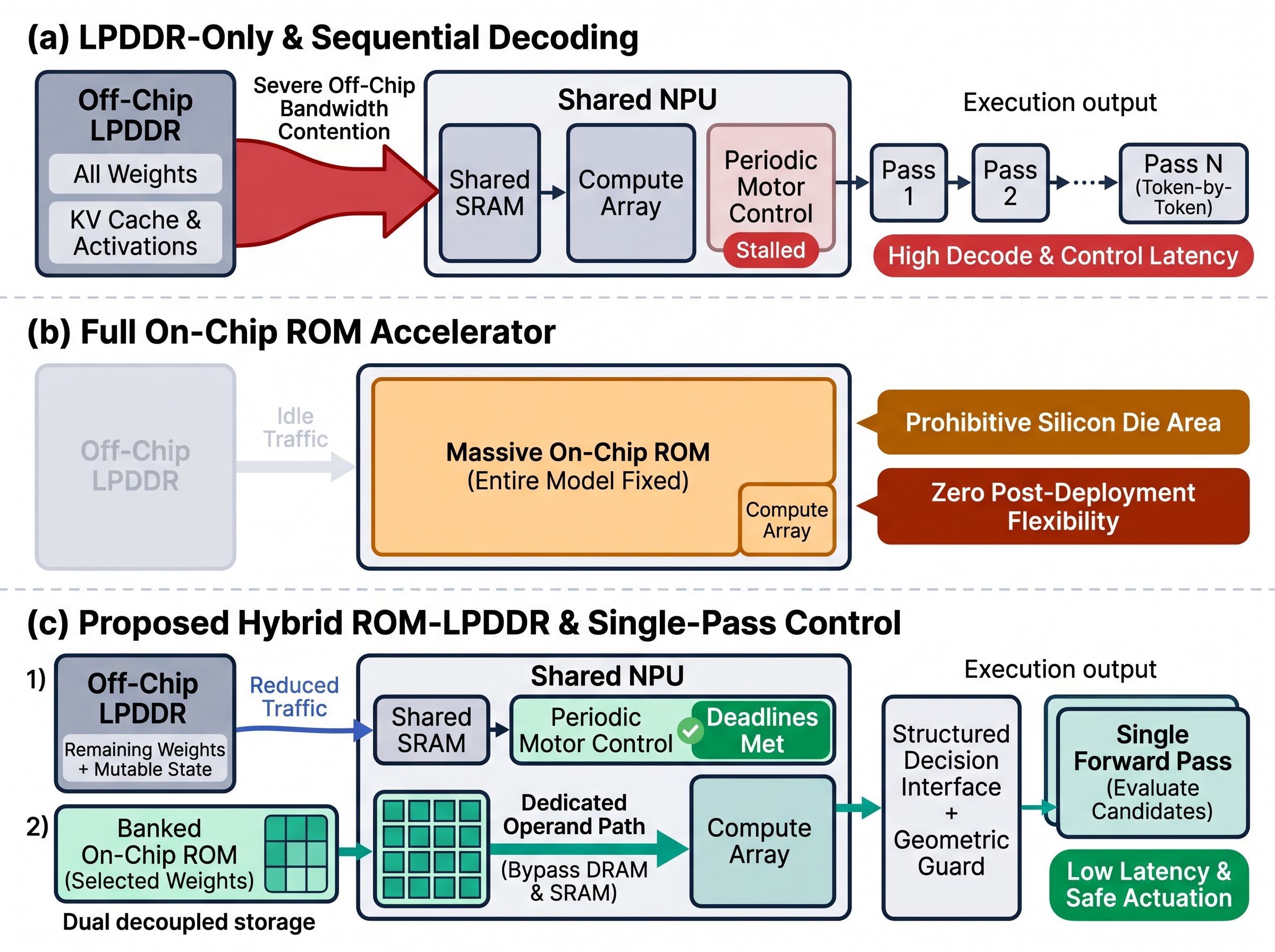}
\vspace{-5pt}
\caption{Edge robot execution architectures. (a) LPDDR-only storage and sequential decoding saturate off-chip bandwidth and stall shared control. (b) Full-model ROM incurs prohibitive die area and freezes base weights. (c) Our hybrid architecture streams fixed weights via banked ROM to an unmodified array while decoupling periodic actuation via a single-pass guarded interface.}
\label{fig:concept}
\vspace{-6pt}
\end{figure}

Conventional architectures store all model parameters in off-chip LPDDR, as shown in Fig.~\ref{fig:concept}(a). Limited package I/O pins and long DRAM access latency keep external bandwidth much lower than on-chip compute throughput~\cite{roofline}. Because single-token decoding offers almost no weight reuse, repeated weight reads across narrow external channels leave compute units idle~\cite{cent}. Quantization~\cite{smoothquant,awq,olive}, parameter-efficient methods~\cite{edgellm}, and hybrid memory offloading~\cite{powerinfer} reduce DRAM capacity pressure, but base weights still cross external interfaces during inference while low-bit formats add runtime dequantization latency~\cite{ant}. To eliminate external weight movement, fixed-weight and ROM-based accelerators~\cite{roma,tom} store fixed weights on-chip (Fig.~\ref{fig:concept}(b)). By exploiting parallel on-chip ROM access, these designs provide high weight bandwidth and reduce data movement energy~\cite{roma,tom}. However, the projected full-model ROM footprints of the larger evaluated LLMs exceed the single-reticle area limit. Fitting an entire model on-chip requires targeting smaller networks or pairing specialized formats, such as ternary weights, with custom compute circuits rather than standard compute arrays. Supporting large models on standard arrays therefore motivates a hybrid ROM-LPDDR memory architecture, placing selected fixed weights in on-chip ROM while keeping remaining data in LPDDR.

\begin{figure*}[!t]
    \centering
    \includegraphics[width=\textwidth]{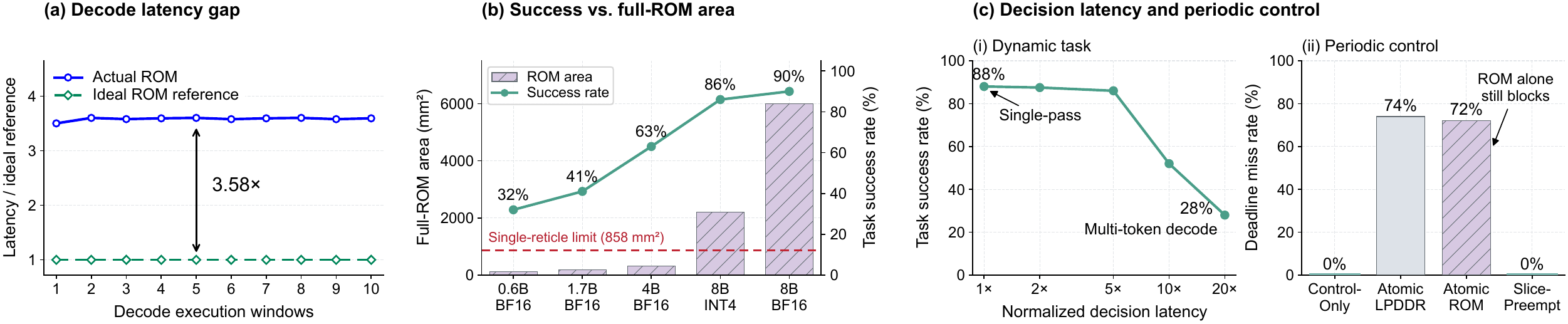}
    \caption{(a) Decode latency comparison between the analytical peak-bandwidth bound (ideal)~\cite{llmcompass} and cycle-level execution under a standard NPU stack (actual)~\cite{llmnpu}, showing up to a $3.58\times$ latency gap.(b) Manipulation task completion rates across Qwen3 scales~\cite{qwen,awq} in ManiSkill3~\cite{maniskill} versus projected full-model planar ROM area~\cite{roma,tom}, with an $858~\mathrm{mm}^2$ single-reticle reference limit. (ci) Dynamic task completion rates under normalized decision latency from single-pass phase scoring ($1\times$) to multi-token decoding ($20\times$). (cii) Periodic motor control deadline miss rates on a shared NPU across memory and scheduling modes.}
    \label{fig:motivation_abc}
\end{figure*}

In practice, hybrid memory hardware alone is insufficient for real-time edge robots, presenting three coupled hardware-software challenges. First, achieving high inference performance requires software mapping and scheduling tailored to the memory hierarchy~\cite{llmcompass}. In Figure~\ref{fig:motivation_abc}(a), the ideal curve estimates lower-bound decode latency from analytical peak-bandwidth models assuming conflict-free overlap~\cite{llmcompass}, whereas the actual curve measures cycle-level execution under a standard NPU software stack~\cite{llmnpu}. Comparing the two curves reveals up to a $3.58\times$ decode latency gap. Naive weight placement wastes scarce ROM capacity on small SRAM-cached tensors, compact low-bit packing causes severe bank conflicts, and uncoordinated ROM-LPDDR reads stall the compute array. Second, preempting a hybrid kernel mid-execution requires saving active accumulators and in-flight ROM requests~\cite{prema}, but running long kernels to completion blocks periodic control on shared hardware~\cite{v10}. This makes preemption granularity a key design choice. Third, because remaining weights and dynamic KV cache stay in LPDDR, multi-step token decoding for discrete robot actions triggers repeated external accesses and inflates decision latency. For choices among predefined actions, evaluating candidates directly offers a way to avoid this sequential generation overhead.

To address these challenges, we propose a hybrid ROM-LPDDR memory architecture co-designed with a heterogeneous scheduling algorithm for edge robots, as shown in Fig.~\ref{fig:concept}(c). At the memory level, the proposed design places large, regular transformer matrices into banked on-chip ROM and supplies a standard compute array via a dedicated ROM datapath. Bitwise XOR address mapping and a word reuse buffer reduce low-bit bank conflicts, and credit-based flow control coordinates fast ROM reads with variable-latency LPDDR accesses without buffer overflow. At the scheduling level, the proposed design arbitrates heterogeneous workloads and aligns preemption with output-slice boundaries, allowing active accumulators and ROM pipelines to drain naturally before periodic control runs. At the decision level, the proposed design decouples slow planning from real-time actuation via a guarded interface with a Jev-style structured decision model~\cite{jev} (hereafter Jev), scoring candidate phases in one forward pass to avoid sequential LPDDR decoding.

We evaluate the proposed design using a cycle-level simulator modeling the compute array, banked ROM, SRAM arbitration, and LPDDR timing across LLM planning and real-time control workloads. On Qwen3-8B, the proposed design reduces the latency of prefill followed by seven decode steps by 34.17\% and achieves a peak decode speedup of 1.56$\times$ over a baseline with all weights in LPDDR. The ROM datapath accelerates weight delivery, while slice-level scheduling prevents long LLM kernels from blocking periodic control. Task response time at P99 falls by 99.56\%, or a factor of approximately $227.27\times$. Our main contributions are as follows:
\begin{itemize}
    \item \textbf{Hybrid ROM-LPDDR Memory Architecture.} We design a hybrid ROM-LPDDR memory architecture with a dedicated ROM datapath and bitwise XOR address mapping to relieve external bandwidth and low-bit bank conflicts.
    \item \textbf{Slice-Level Computation Splitting.} We decompose long inference kernels at output-slice boundaries so on-chip pipelines drain naturally, and replace sequential action decoding with single-pass phase scoring.
    \item \textbf{Heterogeneous Real-Time Scheduling.} We design a scheduler with credit-based flow control that coordinates fast ROM and variable-latency LPDDR operand delivery, prioritizing periodic control tasks without state-saving overhead.
\end{itemize}

\section{Background and Motivation}
\label{sec:background}
\label{sec:preliminary}

\subsection{Edge LLM Memory Architectures}
Autoregressive LLM decoding streams static weights with minimal reuse alongside dynamic KV cache~\cite{pagedattention,attacc}, creating a severe bandwidth bottleneck~\cite{roofline,llmcompass,neupims,cent}. Compression~\cite{smoothquant,awq,qlora,edgellm,gobo,spatten}, sparse offloading~\cite{powerinfer}, and low-bit accelerators~\cite{tender,bitmod,mant,bitfusion} reduce traffic on edge NPUs~\cite{llmnpu}, but weights still cross narrow DRAM channels at every step. Pushing beyond external DRAM, the hardware frontier stores fixed LLM weights directly in on-chip ROM~\cite{roma,tom}. Most notably, TOM~\cite{tom} synthesizes ternary weights into sparsity-aware standard-cell ROM co-located with compute lanes and dynamic power gating. This co-design removes off-chip weight transfers and delivers high storage density and energy efficiency. However, single-die area limits restrict TOM to small models and bind ternary weights to custom circuits rather than standard arrays. Consequently, current architectures remain divided between bandwidth-bound DRAM execution and area-limited full-ROM designs, leaving open hybrid memory and heterogeneous scheduling co-design for billion-parameter models on standard hardware.

\subsection{Structured Decision Models}
\label{sec:background_jev}
Robot decisions can be formulated as selection among available skills: SayCan combines LLM scores with skill affordances~\cite{saycan}, while RT-2 represents actions as generated text tokens~\cite{rt2}. Jev is a structured decision model introduced by TypeSafe AI. It maps unstructured inputs to predefined output types and returns decisions with probabilities through parallel output generation~\cite{jev}. This formulation fits discrete robot choices, such as selecting the next manipulation phase, without generating a textual action description token by token.

\subsection{Embodied Execution and Shared-NPU Scheduling}
In robotics, edge execution is shifting from small control policies~\cite{ppo} on specialized SoCs~\cite{roboshape,rose,robox} toward general embodied foundation models~\cite{saycan,cap,rt2,openvla}. Because cloud offloading~\cite{fogros} incurs link delays and token-based actions require sequential passes, recent studies explore embodied hardware co-design~\cite{daducorki,robomorphic} and single-pass phase scoring~\cite{jev}. At this frontier, DaDu-Corki~\cite{daducorki} predicts multi-step action trajectories to lower LLM invocation frequency and accelerates trajectory-to-torque conversion via a pipelined hardware unit. This decoupled pipeline hides communication latency and improves manipulation speed over frame-by-frame execution. However, DaDu-Corki still offloads heavy LLM inference to remote servers and relies on dedicated control hardware rather than a shared edge NPU. Meanwhile, existing NPU schedulers target wafer-scale servers~\cite{wscllm}, manage caches without hard deadlines~\cite{camdn}, or checkpoint mid-kernel state to DRAM at high preemption cost~\cite{prema}. Thus, an important frontier gap remains: coordinating dual-source ROM-LPDDR delivery with output-slice preemption and single-pass decisions alongside real-time control on one edge NPU.

\begin{figure*}[!t]
    \vspace{-2pt}
    \centering\includegraphics[width=0.97\textwidth]{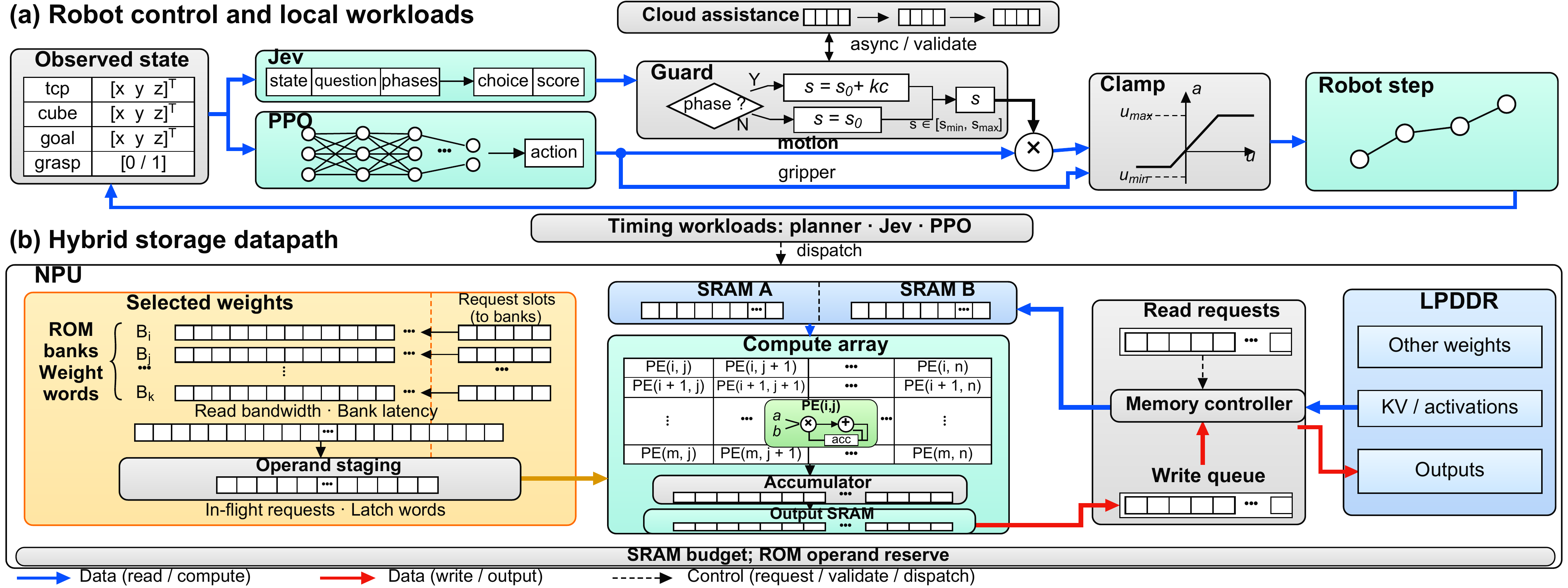}
    \vspace{-5pt}
    \caption{Schematic of the system architecture and hybrid memory datapath organization. (a) Inside the control loop, Jev scores predefined task phases in one forward pass, while a geometric guard modulates motor commands and filters asynchronous remote replies. (b) On the shared NPU, selected fixed weights stream from banked digital ROM through bounded staging directly into the systolic array, bypassing LPDDR queues and single-port SRAM partitions.}
    \label{fig:workflow}
    \vspace{-6pt}
\end{figure*}

\subsection{Task Quality Limited by Memory Size}
\label{sec:mot_memory}
To quantify how model and die size constrain task completion quality, we evaluate Qwen3 planners from 0.6B to 8B~\cite{qwen} (with INT4 compressed via AWQ~\cite{awq}) on robotic manipulation tasks in ManiSkill3~\cite{maniskill}. Full-model silicon footprints are projected from reported planar digital ROM densities~\cite{roma,tom} against the $858~\mathrm{mm}^2$ single-reticle reference limit in Fig.~\ref{fig:motivation_abc}(b). As shown in Fig.~\ref{fig:motivation_abc}(b), compact 0.6B and 1.7B models have smaller ROM footprints, but achieve only $32\%$ and $41\%$ task completion rates due to limited reasoning capacity. Scaling to 4B raises the task completion rate to $63\%$ and already exceeds the single-reticle area limit. Reliable task completion requires the 8B model ($86\%$ in INT4 and $90\%$ in BF16), yet storing all weights in on-chip ROM demands over $2{,}000~\mathrm{mm}^2$ and $5{,}000~\mathrm{mm}^2$ of silicon, respectively. Thus, neither small full-ROM models nor bandwidth-bound DRAM execution suffices, motivating hybrid memory and heterogeneous scheduling co-design.

\subsection{Control Loops Degraded by Inference Latency}
\label{sec:mot_latency}
Following closed-loop SoC-physics evaluation~\cite{rose}, we inject normalized decision delays into dynamic ManiSkill3 tasks~\cite{maniskill} to compare single-pass phase scoring ($1\times$~\cite{jev}) against multi-token autoregressive decoding ($10\times$--$20\times$~\cite{rt2,openvla}). In Fig.~\ref{fig:motivation_abc}(ci), task completion rate holds near $88\%$ at $1\times$--$5\times$ latency, but drops sharply to $52\%$ at $10\times$ and $28\%$ at $20\times$ because physical states evolve before multi-step decoding finishes. Beyond decision staleness, Jev requires a responsive fallback when its decisions are pending or fail the geometric checks. A periodic controller based on proximal policy optimization (PPO)~\cite{ppo} provides this fallback actuation. We profile its execution alongside Qwen3-8B inference on a shared NPU using cycle-level systolic~\cite{scalesim} and LPDDR5~\cite{ramulator2} simulation. Figure~\ref{fig:motivation_abc}(cii) shows that non-preemptive execution (\emph{Atomic LPDDR}) misses $74\%$ of periodic control deadlines. Crucially, adding faster ROM hardware alone (\emph{Atomic ROM}) still misses $72\%$ of deadlines because long inference kernels monopolize the compute array~\cite{prema,camdn}. Restoring a $0\%$ miss rate (\emph{Slice-Preempt}) requires pairing hybrid memory with output-slice preemption and single-pass decisions.

\section{Hybrid Memory and Heterogeneous Scheduling Co-Design}
\label{sec:methodology}

Running billion-parameter planners alongside periodic motor control on a shared edge NPU requires co-designing memory delivery, kernel scheduling, and decision execution. Figure~\ref{fig:workflow} shows the overall proposed architecture built around a standard systolic array. At the memory level, banked digital ROM supplies selected fixed weights through a dedicated ROM datapath with credit-based flow control and conflict-aware low-bit address mapping (Section~\ref{sec:rom}). At the scheduling level, the scheduler arbitrates heterogeneous workloads and restricts preemption to output-slice boundaries where on-chip state drains naturally (Section~\ref{sec:integration}). At the decision level, the proposed design decouples slow planning from real-time actuation via single-pass phase scoring and a non-blocking geometric guard (Section~\ref{sec:decision}).

\subsection{Hybrid ROM-LPDDR Memory Architecture}
\label{sec:rom}
\label{sec:placement}
\label{sec:quantization}

We place selected fixed weights in ROM and retain the remaining weights and mutable state in LPDDR, with SRAM buffering active data. This separation reduces competition for external memory bandwidth while preserving writable storage for model adaptation and runtime state.

\begin{figure}[!t]
  \vspace{-2pt}
  \centering\includegraphics[width=0.92\linewidth]{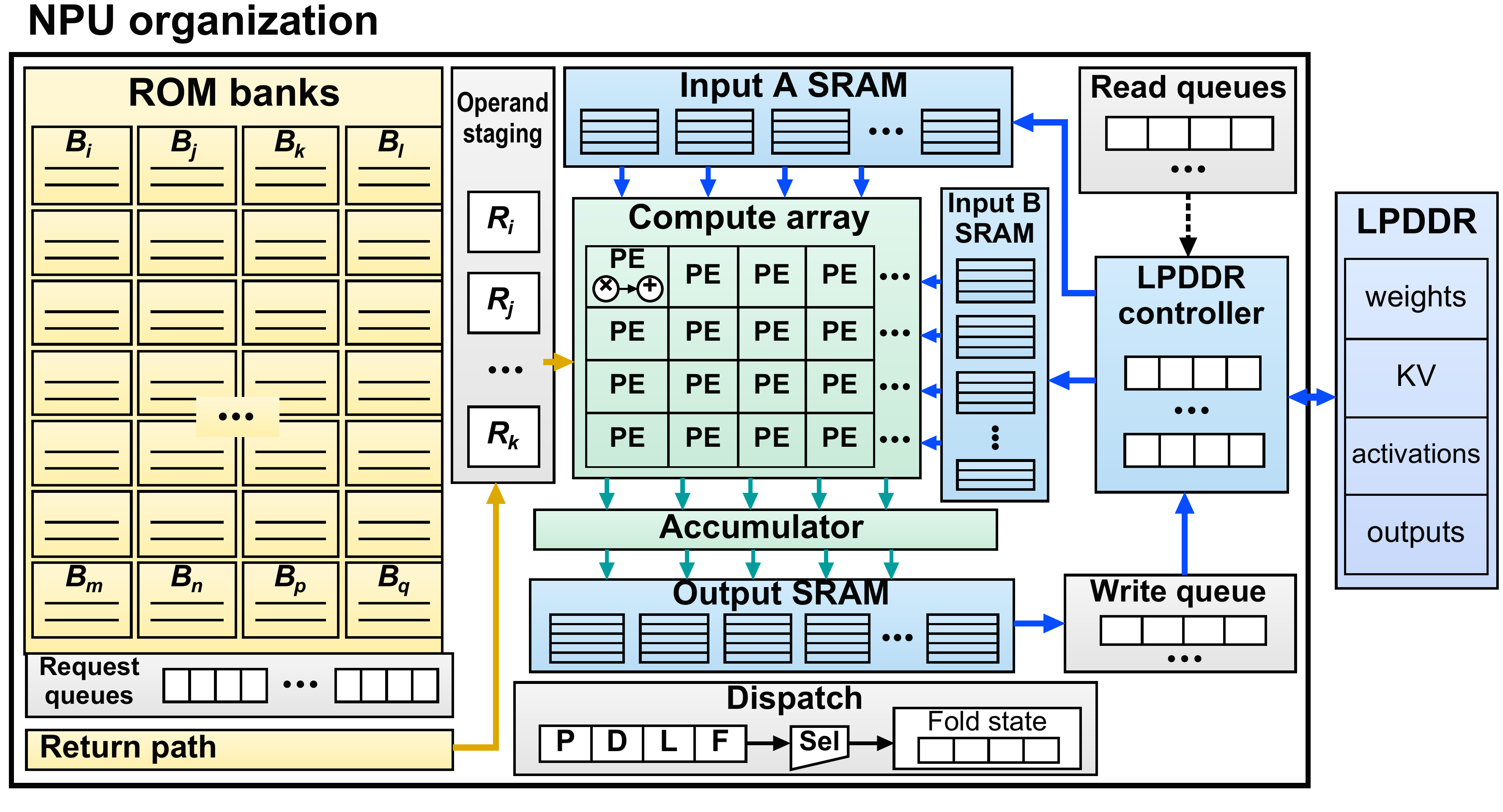}
  \vspace{-5pt}
  \caption{NPU datapath schematic. Dedicated ROM staging and banked input SRAM feed the compute array in lockstep; output SRAM drains to LPDDR.}
  \label{fig:npu_placement}
  \vspace{-7pt}
\end{figure}

\textbf{Dedicated ROM Datapath.}
In conventional NPUs, streaming weights from LPDDR contends for both external channels and on-chip SRAM ports. Baseline scratchpads organize input and output partitions into single-port SRAM banks. Refilling a cache line from LPDDR occupies banks across the target input partition for a full cycle, stalling the lockstep array whenever demand reads collide with active refills. Unlike compute-in-ROM macros that modify array circuits~\cite{roma,tom}, the proposed design adopts standard banked digital ROM and isolates the weight interface (Fig.~\ref{fig:npu_placement}). Selected fixed weights bypass LPDDR queues, line refills, and SRAM input ports to feed the unmodified array via dedicated staging latches, leaving LPDDR and banked SRAM to serve activations, remaining weights, and dynamic KV cache. Across $B$ interleaved ROM banks of word byte width $s_w$, byte address $a$ maps to bank $\lfloor a/s_w\rfloor\bmod B$. Identical word requests within one array beat coalesce into a single bank read and broadcast across consumer lanes. With bank initiation interval $\delta$, global issue width $I$, return width $O$, in-flight limit $F$, and bank latency $L$, steady-state ROM throughput without reuse satisfies:
\begin{equation}
\lambda_{\mathrm{ROM}}\leq\min\{B/\delta,\ I,\ O,\ F/L\}.
\label{eq:supply}
\end{equation}

\textbf{Credit-Based Flow Control.}
Pairing fast, deterministic ROM with variable-latency LPDDR on a lockstep array risks buffer overrun if ROM prefetching continues unchecked during LPDDR misses. The ROM controller bounds outstanding requests and staging occupancy through dual-threshold credit admission (Algorithm~1). Admitting an array beat within lookahead window $W_{\mathrm{win}}$ atomically reserves staging latches for its deduplicated words up to capacity $C_{\mathrm{stage}}$. At each clock edge, ready FIFO returns retire into pre-reserved slots before the round-robin arbiter issues new bank reads, exposing freed capacity without combinational loops. Staging credits release only when ROM words, SRAM operands, and format conversions jointly advance the array.

\textbf{Selective Placement and Low-Bit Mapping.}
Under scarce ROM area, small parameter tensors are already hidden by SRAM prefetching. The proposed design places complete MLP down-projection matrices from transformer layers into aligned ROM blocks within capacity $C_{\mathrm{ROM}}$. These large matrices execute in both prefill and decode, exceed SRAM capacity, present uniform strides across layers, and keep quantization groups within one memory space. For compact low-bit storage, the proposed design packs quantized weights into $s_w$-byte words in groups of size $G$ sharing a floating-point scale $s_g$, mapping each logical operand to one packed word plus one scale word. Because sub-word packing shrinks row strides and concentrates strided reads onto a few banks, for $B=2^m$ banks and cache-line size $L_{\mathrm{line}}$, the proposed design applies a single-cycle intra-line XOR mapping within each line $\ell=\lfloor a/L_{\mathrm{line}}\rfloor$ from base bank $u=\lfloor a/s_w\rfloor\bmod B$:
\begin{equation}
\begin{split}
k&=\bigl(\ell\oplus(\ell\gg m)\oplus(\ell\gg 2m)\bigr)\bmod B,\\
a'&=L_{\mathrm{line}}\ell+s_w(u\oplus k).
\end{split}
\label{eq:interleave}
\end{equation}
Folding $m$-bit line segments distributes power-of-two strides across all $B$ banks without altering line boundaries or image size. A $B$-bank direct-mapped reuse buffer then captures intra-group word and scale locality in one cycle, feeding a $P$-lane converter ($\widehat{w}_j=s_g q_j$) before array execution.

\subsection{Heterogeneous Preemptive Scheduling}
\label{sec:integration}
We divide long inference kernels into output slices and prioritize urgent control requests at slice boundaries. This limits blocking by the planner without the state preservation required for arbitrary interruption~\cite{prema}.

To achieve output-slice preemption, the proposed design decomposes an $(M,K)\times(K,N)$ matrix multiplication on the $A\times A$ array into $\lceil M/A\rceil\lceil N/A\rceil$ complete output folds, where $A$ is the number of processing elements along each array dimension. Preemption is restricted to output-slice boundaries comprising $N_{\mathrm{slice}}$ beats across a configurable number of complete folds. At each boundary, accumulators write back to output SRAM and all in-flight ROM queues and staging latches drain naturally to zero, eliminating context-saving traffic.

Across heterogeneous robot workloads, the proposed design schedules ready jobs by strict stage criticality ($\mathrm{control}>\mathrm{decode}>\mathrm{decision}>\mathrm{prefill}$) and earliest deadline within each class. Compact motor control and single-pass decision kernels execute atomically to minimize dispatch delay, while long prefill and decode kernels yield cooperatively at slice boundaries upon higher-priority arrivals. Context-switch overhead occurs only when switching between distinct jobs, preserving array throughput alongside hard periodic deadlines.

\begin{figure}[!t]
  \vspace{-2pt}
  \hrule height 0.8pt
  \vspace{1.5pt}
  \noindent\textbf{Algorithm 1:} Credit-Based ROM Operand Service (Per Slice)
  \vspace{1.5pt}
  \hrule height 0.4pt
  \vspace{1pt}
  {\footnotesize
  \begin{algorithmic}[1]
  \STATE Initialize admission/retirement pointers, credits, and queues
  \WHILE{the current slice has unretired beats}
    \WHILE{the next beat is within the slice and $W_{\mathrm{win}}$, and its unique words fit free staging}
      \STATE Deduplicate words; atomically reserve staging credits
      \STATE Enqueue mapped bank reads; advance admission pointer
    \ENDWHILE
    \STATE Retire up to $O$ ready returns into reserved staging slots
    \STATE Issue up to $I$ reads round-robin from ready banks, subject to $F$ outstanding requests and bank interval $\delta$
    \STATE Tag issued reads with return time $t+L$
    \IF{the next beat's ROM words, SRAM operands, and conversions are ready}
      \STATE Advance the array and retirement pointer; release beat credits
    \ENDIF
    \STATE $t\gets t+1$
  \ENDWHILE
  \end{algorithmic}
  }
  \vspace{1pt}
  \hrule height 0.8pt
  \vspace{-7pt}
\end{figure}

\subsection{Decoupled Single-Pass Decision Execution}
\label{sec:decision}
We integrate Jev into the robot control flow to score candidate phases in one forward pass, while keeping planning asynchronous and validating decisions before actuation. This avoids sequential token generation for routine choices and allows control to continue while decisions are pending.

For local phase selection, we adapt Jev as a lightweight bidirectional transformer~\cite{jev} that encodes task instructions, robot state, and all candidate phase descriptions into one sequence $x$. A typed attention head refines representations with segment embeddings $e_{\mathrm{type}}$, and a shared scorer $f_\theta$ extracts logits $z_k=f_\theta(h_{m_k})$ at each candidate marker $m_k$:
\begin{equation}
h=H_\theta(E_\theta(x)+e_{\mathrm{type}}),\quad
p_k=\exp(z_k/T)\big/\textstyle\sum_j\exp(z_j/T),
\label{eq:decision}
\end{equation}
where $T$ is the temperature. One forward pass evaluates all candidates concurrently to select $\arg\max_k p_k$ across predefined manipulation phases.

A PPO controller~\cite{ppo} provides fallback actuation when Jev decisions are pending or fail validation. A deterministic geometric guard checks the selected phase against gripper-to-object distances and grasp state at every control step. Terminal or fault phases zero translational velocity immediately. In active phases, kinematic agreement scales the underlying controller velocity commands by $\alpha+(1-\alpha)c$ using normalized confidence $c\in[0,1]$ and conservative floor $\alpha\in(0,1)$, whereas geometric conflicts or pending remote updates fall back to $\alpha$ without scaling gripper closure. Low confidence or phase conflict triggers an asynchronous planner request without stalling periodic control; returned plans apply only if their identifier, validity window, and target phase match the current physical state.

\section{Evaluation}
\label{sec:experiments}

\subsection{Experimental Setup}
\begin{figure}[!t]
\centering
\includegraphics[width=\columnwidth]{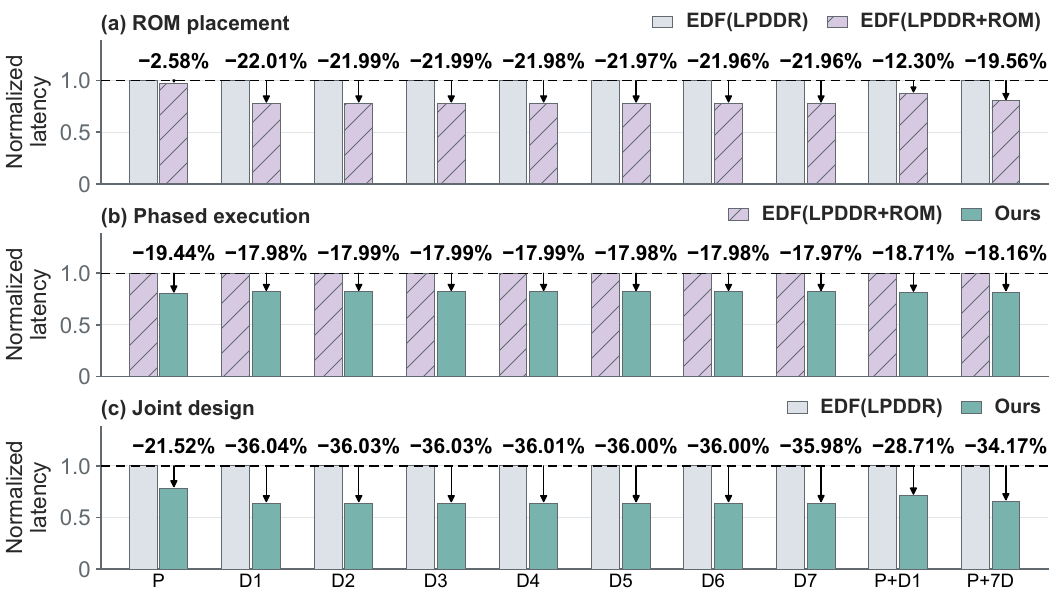}
\caption{Qwen3-8B latency across prefill (P) and seven decode stages (D1--D7). Each panel normalizes its reference to one: (a) EDF (LPDDR+ROM) versus EDF (LPDDR), (b) Ours versus EDF (LPDDR+ROM), and (c) Ours versus EDF (LPDDR). P+7D covers the displayed planning sequence.}
\label{fig:qwen8_phases}
\end{figure}

\textbf{Computing environment.}
CPU nodes with Intel Xeon Gold 6230 and 5320 processors run cycle simulation; NVIDIA A800 and V100 GPUs support model execution and ManiSkill3 robot simulation. Reported NPU inference latency comes from cycle simulation rather than host GPU timing.

\textbf{System simulator.}
We extend SCALE-Sim~\cite{scalesim} and Ramulator~2.0~\cite{ramulator2} with banked ROM, SRAM arbitration, operand staging, and heterogeneous scheduling. A shared cycle timeline captures array stalls, memory contention, and queue backpressure. ManiSkill3~\cite{maniskill} models robot tasks, and ns-3~\cite{ns3} provides network replay.

\textbf{Model and baselines.}
The main workload combines Qwen3-8B~\cite{qwen} prefill and decode with Jev decisions and periodic control. We use Laya~\cite{laya} as the local decision implementation, adapting its inputs to robot state and candidate phases and integrating its outputs with geometric validation and asynchronous planning. PPO provides fallback actuation when Jev decisions are pending or fail validation. Earliest deadline first (EDF) selects the ready request with the earliest deadline at each kernel boundary. EDF (LPDDR) keeps all weights in LPDDR; EDF (LPDDR+ROM) adds selective ROM placement without changing scheduling. Ours combines ROM with phase-specific scheduling and lookahead. Table~\ref{tab:eval_baselines} lists the shared configuration and differences among schemes.

\begin{table}[tb]
\centering\footnotesize
\caption{Qwen3-8B evaluation settings. P/D denotes prefill/decode. ROM is enabled only in EDF (LPDDR+ROM) and Ours; all schemes reserve the same operand-buffer resources.}
\label{tab:eval_baselines}
\setlength{\tabcolsep}{3pt}
\renewcommand{\arraystretch}{1.12}
\begin{tabular*}{\columnwidth}{@{\extracolsep{\fill}}ll@{}}
\toprule
\multicolumn{2}{@{}l}{\textbf{Common hardware configuration}}\\
\midrule
Compute array / clock & $128\times128$ / 1 GHz\\
Total SRAM & 8 MiB\\
LPDDR capacity & 24 GiB\\
LPDDR peak bandwidth & 102.40 GB/s\\
Weight precision & BF16\\
Prefill length & 128 tokens\\
Operand-buffer reservation & 512 KiB within total SRAM\\
In-flight capacity / latch words & 4K / 8K\\
\midrule
\multicolumn{2}{@{}l}{\textbf{ROM configuration (EDF (LPDDR+ROM) and Ours)}}\\
\midrule
Capacity budget & 2.74 GiB\\
Weight placement & 29 complete MLP down projections\\
Bank count / read width per bank & 64 / 64 bits\\
Peak bandwidth & 512 GB/s\\
Read latency & 12 cycles\\
\bottomrule
\end{tabular*}
\vspace{3pt}

\begin{tabular*}{\columnwidth}{@{\extracolsep{\fill}}llll@{}}
\toprule
\multicolumn{4}{@{}l}{\textbf{Scheduling and execution differences}}\\
\midrule
Setting & \shortstack{EDF\\(LPDDR)} & \shortstack{EDF\\(LPDDR+ROM)} & Ours\\
\midrule
Scheduling & \shortstack[l]{Deadline\\order} & \shortstack[l]{Deadline\\order} & \shortstack[l]{Stage priority\\+ deadline order}\\
Preemption & Kernel end & Kernel end & Output slice\\
Folds/slice (P/D) & Whole kernel & Whole kernel & 4 / 16\\
Lookahead (P/D) & 16 / 16 & 16 / 16 & 32 / 16\\
\bottomrule
\end{tabular*}
\par\vspace{2pt}
\raggedright Lookahead is measured in rows. EDF (LPDDR) stores all weights in LPDDR.
\end{table}

\subsection{Performance and Sensitivity Analysis}

\textbf{Main inference result.}
Figure~\ref{fig:qwen8_phases}(c) shows that the joint design reduces prefill plus seven decode steps by 34.17\%, corresponding to a \textbf{1.52$\times$ speedup} over EDF (LPDDR). Decode latency falls by 35.98--36.04\% throughout the seven steps, while prefill improves by 21.52\%. The larger gain during decode addresses the central memory challenge: each generated token requires another traversal of the weights, whereas prefill reuses weight tiles across prompt tokens. The stable decode improvement shows that acceleration persists as dependent tokens are generated.

\textbf{Contribution of ROM placement.}
Figure~\ref{fig:qwen8_phases}(a) isolates the dedicated ROM path while retaining the baseline scheduler. ROM placement reduces decode latency by 21.96--22.01\%, compared with only 2.58\% for prefill. Fixed weights reach the array without competing for LPDDR channels or SRAM refill ports. This directly benefits decode, where weight reuse is limited. Prefill amortizes weight transfers across more arithmetic, so removing these transfers has less influence on its total execution time. Selective placement therefore targets the repeated weight service that limits decoding rather than accelerating all phases equally.

\textbf{Contribution of scheduling.}
Figure~\ref{fig:qwen8_phases}(b) measures the additional benefit of phase-specific scheduling and operand lookahead over EDF (LPDDR+ROM). This reduces prefill latency by 19.44\% and decode latency by 17.97--17.99\%. Faster weight supply alone does not determine when ready work can execute: scheduling sets the execution order and slice boundaries, while lookahead controls how early operands are requested. Coordinating these choices converts the dedicated supply path into useful array execution. The joint design reduces P+D1 latency by 28.71\%, rising to 34.17\% for P+7D as the sequence includes more of the strongly accelerated decode phase. Thus, memory placement and execution policy make complementary contributions to the planning sequence.

\begin{figure}[!t]
\centering
\includegraphics[width=\columnwidth]{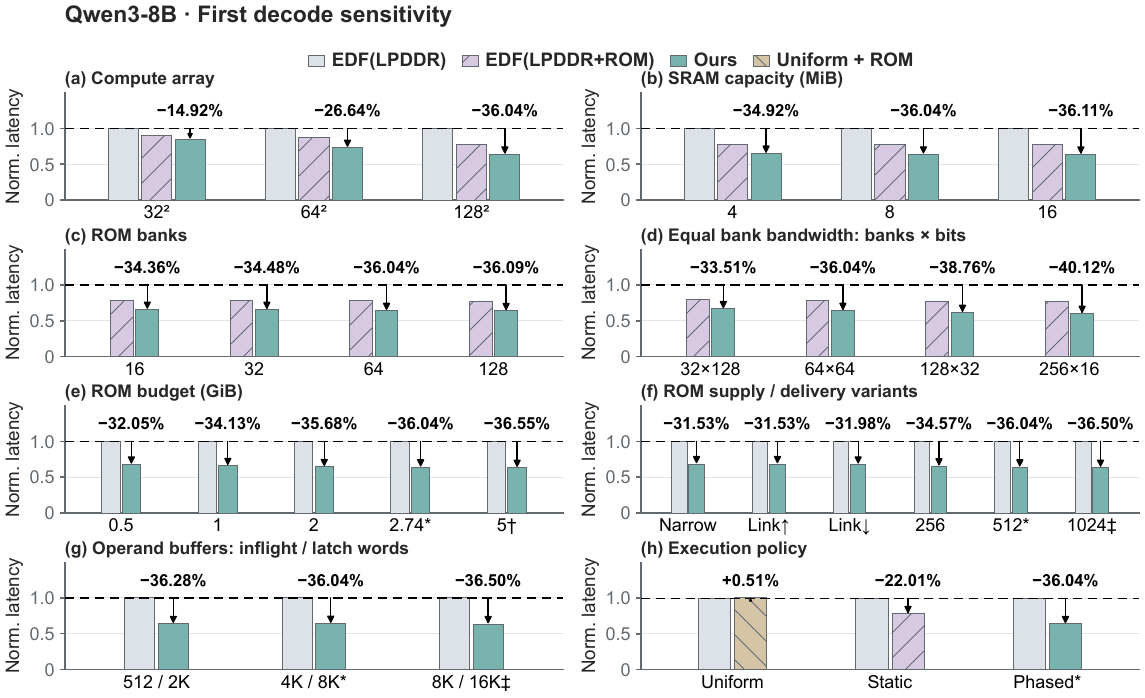}
\caption{Qwen3-8B first-decode sensitivity, normalized to the EDF (LPDDR) reference of each comparison. Panels vary array size, SRAM, ROM banks, bank organization, capacity, delivery, buffers, and execution policy. In (f), 256/512/1024 are ROM peak GB/s. The 1024 point also enlarges interfaces and buffer reservation; the 5~GiB budget stores 3.38~GiB of selected weights.}
\label{fig:qwen8_sensitivity}
\end{figure}

\textbf{Hardware sensitivity.}
Figure~\ref{fig:qwen8_sensitivity} shows that acceleration depends on balanced compute, bank concurrency, and operand delivery rather than capacity alone. Widening a link behind a narrow ROM source provides no further gain, while insufficient in-flight and staging capacity causes a 0.72\% slowdown. Larger buffers also consume SRAM needed by other operands; the largest-buffer point changes interfaces and reservation together. The execution-policy comparison reinforces the main ablation: uniform settings slightly slow execution, whereas phased execution achieves the largest gain.

\textbf{Area feasibility.}
We assess a B-ROM implementation candidate following ROMA~\cite{roma}, retaining the main experiment's ROM capacity and BF16 precision. Assuming a 40.00\% reduction in ROM macro area, scaled SRAM and compute estimates, and a 20.00\% integration allowance, the projected area at 7~nm is $525.80~\mathrm{mm}^2$, a 38.47\% reduction from the standard ROM estimate. Shared address logic and fixed wiring reduce the storage footprint without quantizing the weights. This projected area falls below the $858~\mathrm{mm}^2$ single-reticle reference limit used in our motivation; physical implementation and timing closure at 1~GHz remain unverified.

\subsection{Generality and Scalability}
\begin{table}[tb]
\centering\small
\caption{Additional BF16 model checkpoints. $L$, $d$, and $d_{\mathrm{FFN}}$ denote layer count, hidden width, and MLP intermediate width; Q/KV denotes attention-head counts.}
\label{tab:eval_models}
\setlength{\tabcolsep}{3pt}
\renewcommand{\arraystretch}{1.15}
\begin{tabular*}{\columnwidth}{@{\extracolsep{\fill}}lrrrr@{}}
\toprule
Model & $L$ & $d$ & $d_{\mathrm{FFN}}$ & Q/KV\\\midrule
Mistral-7B-Instruct-v0.3 & 32 & 4096 & 14336 & 32/8\\
Qwen2.5-7B-Instruct & 28 & 3584 & 18944 & 28/4\\
Phi-3.5-mini-instruct & 32 & 3072 & 8192 & 32/32\\
\bottomrule
\end{tabular*}
\end{table}

\begin{figure}[!t]
\centering
\includegraphics[width=\columnwidth]{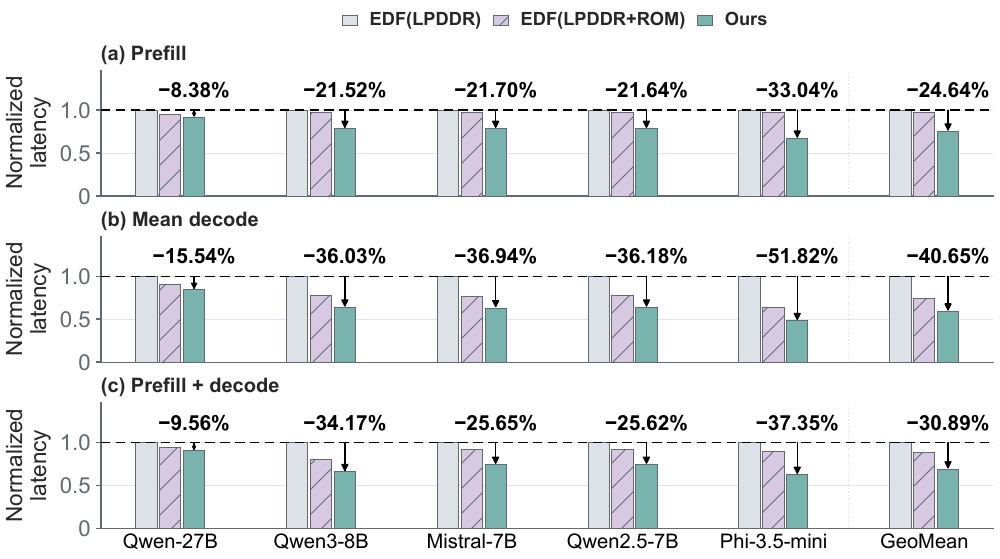}
\caption{Latency across five LLMs under matched hardware budgets: (a) prefill, (b) mean decode over D1--D2 for the four smaller LLMs and D1 for 27B, and (c) prefill plus decode, covering seven decode tokens for Qwen3-8B, two for the other smaller LLMs, and one for 27B. Each LLM is normalized to its EDF (LPDDR) baseline. Labels show the reduction of Ours. GeoMean summarizes the four smaller LLMs over their respective evaluation windows; the 27B result is shown separately.}
\label{fig:model_latency}
\end{figure}

Figure~\ref{fig:model_latency} evaluates the additional BF16 LLMs in Table~\ref{tab:eval_models} and Qwen3.5-27B~\cite{qwen27}. Mean decode latency over D1--D2 falls by 36.18--51.82\% across the three additional smaller LLMs, exceeding their prefill improvements and reproducing the main experiment's benefit from reduced weight streaming. The 27B model retains a 15.54\% reduction for its first decode token. Its smaller gain reflects the reduced fraction of a larger weight footprint served by a fixed ROM budget.

\subsection{Robot Task Performance}
\textbf{Task completion.}
The proposed design successfully completes all 100 object grasping and transport trials in the main experiment, achieving a \textbf{100.00\% task completion rate (100/100)}. Jev guides phase selection, the geometric guard validates decisions, and fallback actuation maintains control when needed. This result measures successful manipulation by the integrated system.

\begin{figure}[!t]
\centering
\includegraphics[width=\columnwidth]{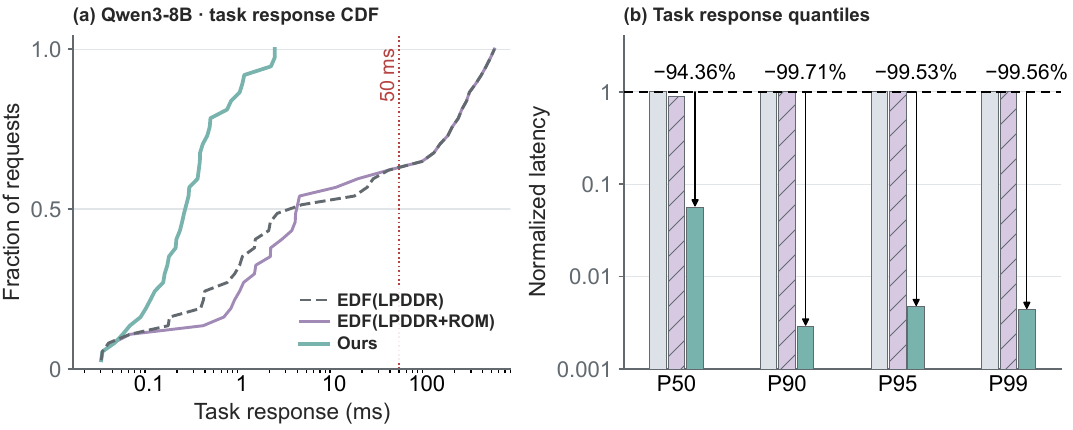}
\caption{Task response time under EDF (LPDDR), EDF (LPDDR+ROM), and Ours. (a) Response time CDF with a 50~ms deadline. (b) Response time quantiles, each normalized to its EDF (LPDDR) value on a logarithmic axis. Labels show the reduction achieved by Ours.}
\label{fig:planning_control}
\label{fig:control_cdf}
\end{figure}

\textbf{Control responsiveness.}
Figure~\ref{fig:planning_control}(a) shows the task response time CDF; all displayed responses of Ours fall below the 50~ms deadline. Figure~\ref{fig:planning_control}(b) compares response time quantiles. Relative to EDF (LPDDR), median response time falls by 94.36\% and P99 by 99.56\%, corresponding to approximately \textbf{227.27$\times$} lower P99 response time.

EDF (LPDDR+ROM) retains long response tails because faster weight delivery does not remove waiting for an active kernel to finish. Output-slice preemption lets ready control requests execute at the next safe boundary. Removing this blocking explains why the response time gain is much larger than the inference speedup: ROM shortens model execution, while scheduling prevents that execution from delaying actuation.

\section{Conclusion}
\label{sec:conclusion}
The proposed design combines hybrid ROM-LPDDR memory, heterogeneous scheduling, and Jev decisions to accelerate edge robotic LLM execution. A dedicated ROM datapath with credit-based flow control reduces weight contention, while output-slice preemption limits control blocking. Evaluating candidate phases in one pass avoids sequential action generation. The joint design reduces planning latency by 34.17\%, delivering a 1.52$\times$ speedup, and lowers P99 task response time by 99.56\%, corresponding to approximately 227.27$\times$ lower latency, with \textbf{100.00\% task completion}. Additional LLMs retain inference gains. These results demonstrate the complementary benefits of selective weight storage, timely compute access, and structured decision execution on a shared edge NPU.

\ifaidisclosure
  \section*{Content Generated by AI}
AI assistance was used for grammar checking of the manuscript. All research ideas, methodological designs, and experimental evaluations are the authors' original work.

\fi

\bibliographystyle{IEEEtran}
\bibliography{ref/style_control,ref/references}
\end{document}